\documentclass[letterpaper, 10 pt, conference]{ieeeconf}  %

\IEEEoverridecommandlockouts                              %

\usepackage{amsmath}
\usepackage{amssymb}
\usepackage{graphicx}
\usepackage{booktabs}
\usepackage{xcolor}
\usepackage[hidelinks]{hyperref}
\definecolor{ucbBlue}{HTML}{3B7EA1}
\definecolor{tblNavy}{HTML}{234E70}
\definecolor{tblHtwo}{HTML}{2B6CB0}
\definecolor{tblDex}{HTML}{805AD5}
\definecolor{tblHocap}{HTML}{B45309}
\definecolor{tblTeal}{HTML}{0F766E}
\definecolor{tblGray}{HTML}{6B7280}
\definecolor{tblBest}{HTML}{2B6CB0}
\definecolor{tblSecond}{HTML}{B7791F}

\title{\LARGE \bf
Grounding Generated Video Plans in Simulation Towards Versatile Dexterous Controllers
}

\author{
Tianyue Wu$^{2,3,1,*,\ddagger}$,
Boyuan An$^{2,1,*}$,
Shuqi Zhao$^{1}$,
Heyu Guo$^{2}$,
Wanli Xing$^{2}$,\\
Yi Ma$^{3,1}$,
Kaifeng Zhang$^{2}$,
Ruihai Wu$^{1,\dagger}$,
and Masayoshi Tomizuka$^{1,\dagger}$
\\[0.4em]
\normalsize
$^{1}$University of California, Berkeley \quad
$^{2}$Sharpa \quad
$^{3}$The University of Hong Kong
\\
\normalsize
$^{*}$Co-first authors \quad
$^{\dagger}$Co-advisors \quad
$^{\ddagger}$Corresponding author
\\[0.3em]
\normalsize
}
\vspace{-0.2cm}

\makeatletter
\long\def\@makecaption#1#2{%
\ifx\@captype\@IEEEtablestring%
\begin{center}{\footnotesize #1}\\{\footnotesize\scshape #2}\end{center}%
\@IEEEtablecaptionsepspace%
\else
\@IEEEfigurecaptionsepspace%
\setbox\@tempboxa\hbox{\footnotesize {\bfseries #1.}~~ #2}%
\ifdim \wd\@tempboxa >\hsize%
\setbox\@tempboxa\hbox{\footnotesize {\bfseries #1.}~~ }%
\parbox[t]{\hsize}{\footnotesize \noindent\unhbox\@tempboxa#2}%
\else%
\ifcenterfigcaptions \hbox to\hsize{\footnotesize\hfil\box\@tempboxa\hfil}%
\else \hbox to\hsize{\footnotesize\box\@tempboxa\hfil}%
\fi\fi\fi}
\makeatother
\newcommand{\figtitle}[1]{\textbf{#1}}

\begin{document}

\makeatletter
\IEEEaftertitletext{%
  \vspace{-1.5em}
  \begin{center}
    \refstepcounter{figure}\label{fig:teaser}%
    \includegraphics[width=\textwidth]{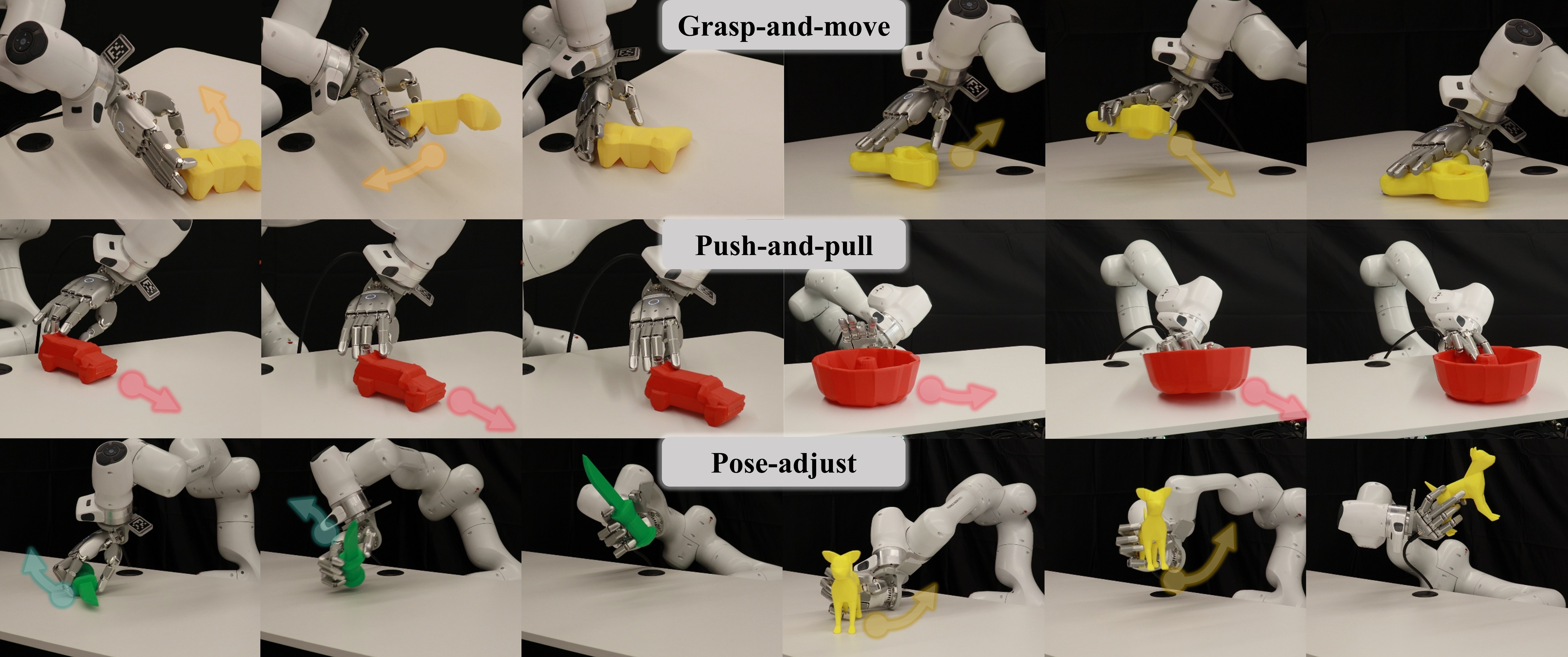}%
    \vskip-0.5em %
\@makecaption{\fnum@figure}{\figtitle{Real-world execution of generated video references.}
    A unified tracking controller performs grasp-and-move, push-and-pull, and
    pose-adjust behaviors across diverse objects.}
  \end{center}
  \vspace{-0.7em}
}
\makeatother

\maketitle
\thispagestyle{empty}
\pagestyle{empty}

\begin{abstract}

Generated hand–object interaction (HOI) videos provide a controllable way to propose manipulation motions.
Simulation-based HOI tracking can translate such kinematic references into feasible 
low-level control, but its scalability is limited by the lack of reliable 
reference motions. We therefore combine generated videos with simulation-based HOI grounding:
during training, generated videos provide diverse motion references for learning a multi-object,
multi-trajectory HOI tracker, and at deployment, the video model produces motion plans that
are executed by the learned tracker. In particular, we propose a method that enables scalable reference
generation by HOI reconstruction with minimal manual intervention and successfully grounds more than 1,500 generated
videos in simulation, achieving success rates over 25 percentage points higher than those of
baselines during simulation-based training.
In real-world closed-loop experiments, it achieves diverse grasps, including functional grasps, 
non-prehensile manipulation, and post-grasp object-pose tracking. Videos and code are available at \href{https://boyuan-an.github.io/GALATEA/}{\textcolor{blue}{this URL}}.

\end{abstract}

\section{INTRODUCTION}

Human children and young animals learn many manipulation skills by watching others.
Although they cannot observe the demonstrator's motor commands, they can see
both how a similar body moves and how those movements affect the surrounding
world.  Together, these visual cues provide dense, indirect supervision for
learning purposeful behavior~\cite{meltzoff1988imitation,whiten2021culture}.
These observations motivate a promising methodology in robotics: dexterous
manipulation skill acquisition from visual demonstrations~\cite{wang2023mimicplay,kareer2024egomimic}.

In-the-wild videos contain hand behaviors at a scale and diversity beyond current
robot demonstration datasets, but robots cannot
interpret them as humans do. The intended interactions are entangled with irrelevant human motion
and uncontrolled variation in viewpoint, appearance, occlusion, and scene context,
so extracting useful references still requires extensive human filtering and
annotation~\cite{paliwal2026doasido}. Prior work obtains cleaner supervision by
recording human manipulation with standardized hardware and prescribed task
protocols~\cite{egoverse}, but collection,
annotation, and validation remain costly. Moreover, videos do not directly
reveal 3-D hand--object states, contacts, forces, or actions executable by a
particular robot embodiment. 
Moving toward versatile dexterous control through these visual demonstrations therefore requires both a more
scalable and controllable source of diverse references and a video retargeter that can
physically ground them.

Accordingly, we study simulation-trained multi-object, multi-trajectory
hand--object tracking as a scalable task \cite{luo2026sonic}, with video generation
models~\cite{teamseedance2026seedance} supplying its training
references. 
Advanced video foundation models
provide reference scale by exposing priors learned from Internet-scale video
corpora through controlled synthesis: image and language conditioning can specify the object,
intended behavior, and scene while suppressing irrelevant variation, producing
diverse and controllable manipulation references with less manual curation compared to 
in-the-wild videos. Modern simulators \cite{makoviychuk2021isaacgym} provide a complementary path to scale by enabling parallel interaction rollouts, which supply dense 
supervision for mapping kinematic hand--object references to dynamically feasible low-level actions.
The resulting general low-level controller can serve as a physics-aware execution layer for
high-level robot foundation models \cite{luo2026sonic}. However, existing video-to-manipulation methods typically reconstruct 
or optimize only a limited set of references \cite{chen2025vividex,lum2025human2sim2robot}, while learning a versatile HOI tracker from 
relatively large, noisy, video-derived reference corpora remains underexplored.

\begin{figure}[t]
  \hspace{-0.3cm}
  \includegraphics[width=1.05\columnwidth]{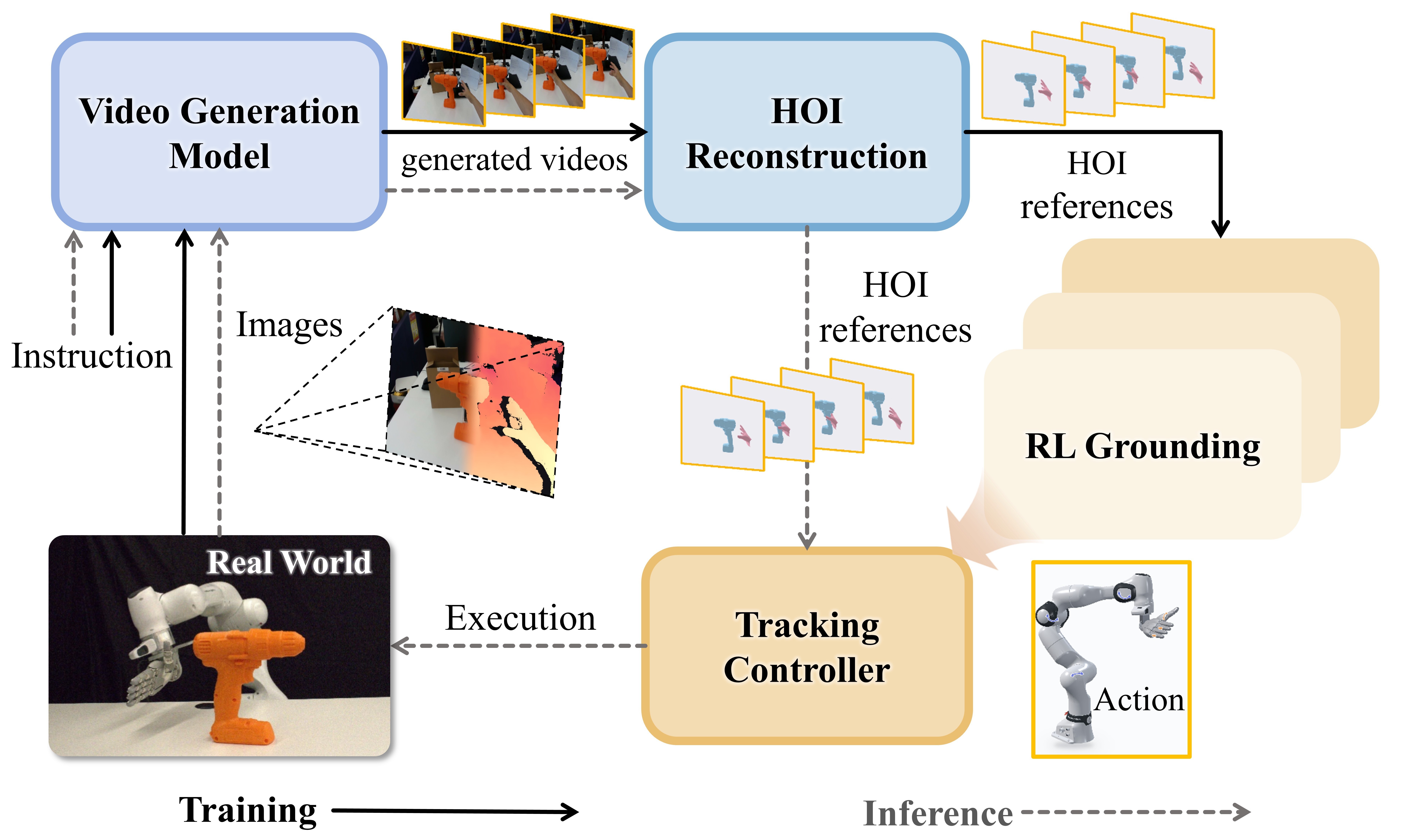}
  \par\vspace{-1.1em}
  \caption{\figtitle{Overview of \texttt{GALATEA}.}}
  \label{fig:galatea_pipeline}
  \vspace{-1em}
\end{figure}

Here, we present \texttt{GALATEA}\footnote{Named after Galatea, the statue brought to life in later retellings of the Pygmalion myth, echoing our goal of turning imagined videos into reality.} to Ground generAted-video pLans At scale with simulation-trained 
Tracking for Executable Actions. During training, generated videos provide diverse motion 
references for learning a multi-object, multi-trajectory tracker; during deployment, prompted
video plans are physically executed by the learned tracker, as shown in Fig.~\ref{fig:galatea_pipeline}. \texttt{GALATEA} consists of two stages. 
First, a reconstruction pipeline combines foundation-model perception, stereo initialization, 
and joint hand--object optimization to recover HOI trajectories from generated
videos. Second, a sim-to-real RL recipe combines a task-agnostic tracking objective,
reference augmentation, and the Split and Aggregate Policy Gradients
(SAPG)~\cite{singla2024sapg} optimizer, outperforming baselines by more than
25 percentage points of success rates. Finally, \texttt{GALATEA} reconstructs about 2,000 usable HOI
trajectories from 2,500 generated clips and more than 1,500 trajectories are successfully grounded in simulation 
 by expert controllers, which are distilled into a unified policy.

Our contributions are summarized as follows:
\begin{itemize}
\item \texttt{GALATEA}, a system that bridges generated HOI video plans and physical execution through HOI reconstruction and a motion tracking controller;
\item a practical HOI reference extraction method from generated videos combining advanced perception model, stereo depth initialization, and joint hand--object optimization;
\item an improved tracking-style RL recipe with design choices in RL formulation and the optimizer to support learning from multiple noisy, video-derived HOI references;
\item closed-loop real-world execution of both trained and unseen video plans,
 which achieves diverse grasps, pushing/pulling, and post-grasp object pose adjustment (Fig.~\ref{fig:teaser}).
\end{itemize}

\section{RELATED WORK}

\vspace{-1.0mm}
\subsection{Generative Video Models for Manipulation}
\vspace{-1.0mm}
Generative video priors enter manipulation policies through several approaches.
Some approaches generate visual futures and translate them into robot actions
with an inverse-dynamics model or a video-conditioned policy
\cite{du2023unipi,bharadhwaj2024gen2act}, whereas others jointly model visual
futures and actions \cite{ye2026dreamzero,li2026lingbot}. Their action
components typically require embodiment-specific robot demonstrations for
training. Methods with an explicit intermediate interface instead convert
generated videos into structured motion references for downstream control
\cite{liang2024dreamitate}.  LVP~\cite{chen2025lvp} retargets generated wrist
and hand kinematics, but does not track object motion or contact during open-loop
execution.  Dex4D~\cite{kuang2026dex4d} conditions a closed-loop,
simulation-trained policy on object-centric 3-D point tracks, but leaves the
desired interaction strategy under-specified, limiting control over functional
grasps.   We instead reconstruct and physically ground both finger-level and object
trajectories, preserving both the demonstrated agent motion and its intended
effect on the object. 
Related work also uses generative video models as motion
planners or data generators for humanoid control
\cite{Genmimic,Imagine2Real,xie2026grail}, rather than for contact-rich dexterous
manipulation. Concurrent works on dexterous manipulation systems that are grounded in the "System 1/System 2" framework \cite{kahneman2011thinking} also employ single-task video imitation policies \cite{gupta2026lucid} or Vision-Language-Action (VLA) models \cite{xing2026decoupled} as motion planners. By integrating a video generation model with fundamentally superior generalization capabilities, we retain the possibility of achieving zero-shot manipulation.

\vspace{-1.5mm}
\subsection{Retargeting from Kinematic Demonstrations for Dexterous Control}
\vspace{-1.5mm}

Kinematic demonstrations provide dense supervision over desired hand and
object motion, while simulation can recover the contact-feasible actions absent
from such references.  Existing approaches can be grouped by the source of
their kinematic supervision.  One line retaregts from high-quality motion capture
or otherwise curated human--object trajectories
\cite{zhao2024dexh2r,li2025maniptrans,zhao2026dexmachina,zhu2026chord,liu2025dextrack,
feng2026regrind,li2026towards}.  A second line reconstructs references from
one or a small number of human videos
\cite{qin2022dexmv,chen2025vividex,hsieh2025dexman,chen2026v2p,lum2025human2sim2robot,chen1dex}, where the trajectories can be noisier than 
those in the first line and are hard to solve using methods like open-loop motion planning.  
A third line, which our work belongs to, 
reduces dependence on manually captured trajectories through synthetic 
or simulation-expanded trajectories \cite{wang2025hophot,adalibieke2026adadextrack,gupta2026lucid}, 
which are more diverse and scalable but introduce additional noise or artifacts.
Across these previous works, only a limited subset using RL to retarget from a 
relatively large set of HOI references \cite{liu2025dextrack,adalibieke2026adadextrack}, 
some of which is limited to simulation dynamics \cite{wang2025hophot}.  
In contrast, we demonstrate that a unified controller can be learned from a
relatively large-scale and noisy corpus of generated-video references, and
empirically establish the superiority of our RL training recipe over baseline methods. 

\section{METHOD}

Our method has two stages. The first generates videos and reconstructs their 3-D
hand--object trajectories. The second uses tracking-style RL and policy distillation
in simulation to learn a unified controller from these trajectories.

\vspace{-1.5mm}
\subsection{Video Generation and Reconstruction}
\vspace{-1.5mm}
\subsubsection{Video Generation from Collected Images}
Minimizing manual effort requires both video generation and reconstruction to
succeed reliably. We therefore use the state-of-the-art proprietary video model
Seedance~2.0~\cite{teamseedance2026seedance}, which produces substantially more
plausible manipulation videos than current open-source alternatives in our
tests. Because simulation-rendered conditioning images often lead to physically
inconsistent interactions, we instead capture a real first-frame RGB image and
condition the model on this image and a language instruction. We consider three 
manipulation types. \emph{Grasp-and-move}
grasps an object from the table, moves it through space with little change in
orientation, and may place it back on the table. \emph{Push-and-pull} moves the
object on the table without lifting. \emph{Pose-adjust} grasps the object and
substantially reorients it in midair. A stereo camera simultaneously captures
first-frame depth, which provides a metric anchor for subsequent reconstruction.

\subsubsection{HOI Reconstruction}

\begin{figure}[t]
  \centering
  \includegraphics[width=\columnwidth]{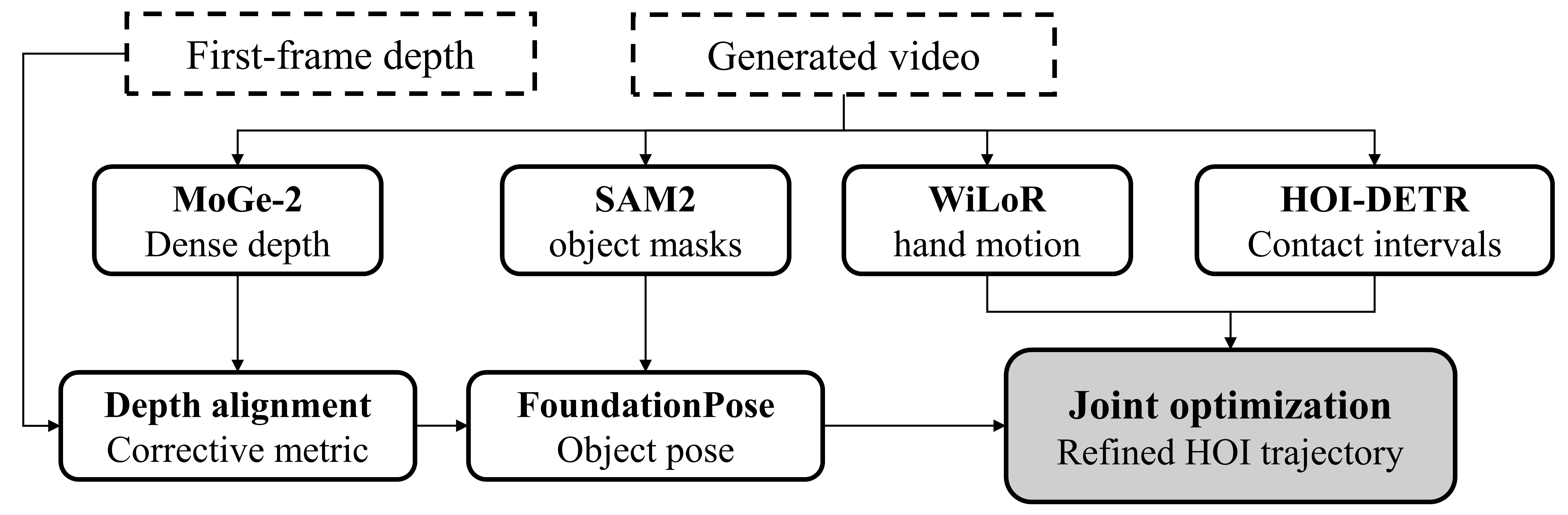}
  \par\vspace{-0.5em}
  \caption{\figtitle{HOI reconstruction pipeline.}  First-frame depth metric-aligns
  generated-video depth. Object masks, initial hand and object motion, and
  contact intervals then guide joint optimization to produce final estimates.}
  \label{fig:hoi_reconstruction_pipeline}
  \vspace{-2.0em}
\end{figure}

We use $H$ and $O$ to denote the hand and object, respectively, and
$X\in\{H,O\}$ to index either entity.
Given a generated video $\{I_t\}_{t=1}^{T}$, calibrated camera intrinsics
$\mathbf K$, a metric object mesh $\mathcal M^O$, and first-frame stereo depth
$D_1^{\mathrm{st}}$, we reconstruct an aligned MANO~\cite{romero2017mano} hand mesh and 6-DoF object trajectory 
by estimating both motions independently and then jointly refining them, inspired by prior
joint hand--object reconstruction methods~\cite{hampali2020honnotate,fan2024hold}, 
which is crucial for the success of downstream policy learning.  Fig.~\ref{fig:hoi_reconstruction_pipeline}
summarizes the pipeline.

\noindent\textbf{Metric Depth Estimation.}
MoGe-2~\cite{wang2025moge2} predicts a dense depth map $\widetilde D_t$, whose
global scale and offset may drift across frames. SAM2~\cite{ravi2024sam2}
provides object masks $M_t^O$ and moving-foreground masks $M_t^F$, with $M_t^F$
covering the hand, forearm, and object. Under the fixed-camera and
static-background assumption, $\Omega_t^{\mathrm{bg}}$ denotes pixels outside
both $M_1^F$ and $M_t^F$, which observe the same background in the first and
current frames.
Following~\cite{xie2026grail}, we use trimmed least squares over
$\Omega_t^{\mathrm{bg}}$ to fit $s_t\widetilde D_t+b_t$ to the first-frame
stereo depth $D_1^{\mathrm{st}}$, and set
$D_t(\mathbf p)=[s_t\widetilde D_t(\mathbf p)+b_t]_+$. This anchors each frame
to a corrective metric scale and offset. $D_t$ is then passed to
object tracking. RANSAC~\cite{fischler1981random} on $D_1^{\mathrm{st}}$ also
defines a gravity-aligned support plane, i.e., the table.

\noindent\textbf{Initial Motion Estimation.}
FoundationPose~\cite{wen2024foundationpose} estimates initial object poses
$\widehat{\mathbf T}_t^O$ from $(I_t,D_t,M_t^O,\mathcal M^O,\mathbf K)$; low mask
coverage triggers a color-adapted retry.  WiLoR~\cite{potamias2025wilor}
estimates MANO vertices $\widehat{\mathbf V}_t^H$ and 2D keypoints $\mathbf u_t$.

\noindent\textbf{Joint Hand--Object Optimization.}
Starting from the initial estimates, we optimize per-frame rigid corrections
$\mathbf T^{O}_{t}=\Delta\mathbf T^{O}_{t}\widehat{\mathbf T}^{O}_{t}$ and
$\mathbf V^{H}_{t}=\Delta\mathbf T^{H}_{t}\widehat{\mathbf V}^{H}_{t}$, while
retaining finger articulation. Removing object and forearm regions from $M_t^F$
gives the observed hand mask $M_t^H$. For $X\in\{O,H\}$, a differentiable
renderer projects the current mesh into a filled 2-D occupancy mask $S_t^X$,
which we call its silhouette. We match it to the observed mask $M_t^X$ using
$\mathcal L_{\mathrm{sil}}^X=\sum_t\|W_t^X\odot(S_t^X-M_t^X)\|_2^2$, where
$W_t^X$ excludes pixels occluded by the other component. Given metric mesh
dimensions and camera intrinsics, the mask position, shape, and size constrain
lateral translation, rotation, and depth. HOI-DETR~\cite{darkhalil2026hoidetr}
detects contact frames $\mathcal T_c$, which gate the contact loss
$\mathcal L_{\mathrm{con}}$.  The full objective is
\vspace{-0.5em}
\begin{equation}
\begin{split}
\mathcal L ={}&
\lambda_{\mathrm{FP}}\mathcal L_{\mathrm{FP}}+
\lambda_{\mathrm{kp}}\mathcal L_{\mathrm{kp}}+
\lambda_{\mathrm{sil}}^{O}\mathcal L_{\mathrm{sil}}^{O} \\
&+\lambda_{\mathrm{sil}}^{H}\mathcal L_{\mathrm{sil}}^{H}+
\lambda_{\mathrm{con}}\mathcal L_{\mathrm{con}}+
\lambda_{\mathrm{temp}}\mathcal L_{\mathrm{temp}}.
\end{split}
\label{eq:hoi_reconstruction}
\vspace{-1.2mm}
\end{equation}
Let $\pi_{\mathbf K}$ denote perspective projection, $\mathbf V^O$ the object
vertices, $\mathbf J_t^H$ the MANO joints, $\mathbf c_t^O$ the object
translation, $\mathbf c_t^H$ the hand-mesh centroid, and $\mathcal C_t$ the
contact-labeled hand-segment vertices.  The remaining losses are
\vspace{-1.2mm}
\begin{equation}
\begin{aligned}
\mathcal L_{\mathrm{FP}}={}&\sum_t\left\|
\pi_{\mathbf K}(\mathbf T_t^O\mathbf V^O)-
\pi_{\mathbf K}(\widehat{\mathbf T}_t^O\mathbf V^O)\right\|_1,\\
\mathcal L_{\mathrm{kp}}={}&\sum_t\left\|
\pi_{\mathbf K}(\mathbf J_t^H)-\mathbf u_t\right\|_1,\\
\mathcal L_{\mathrm{con}}={}&\sum_{t\in\mathcal T_c}\sum_{\mathbf v\in\mathcal C_t}
d^2(\mathbf v,\mathbf T_t^O\mathcal M^O),\\
\mathcal L_{\mathrm{temp}}={}&\sum_{X\in\{O,H\}}\sum_t
\left(\|\Delta\mathbf c_t^X\|_1+
\beta\|\Delta^2\mathbf c_t^X\|_1\right).
\end{aligned}
\label{eq:hoi_loss_terms}
\vspace{-1.2mm}
\end{equation}
Here $d(\cdot,\cdot)$ is point-to-mesh distance and $\Delta$ is the temporal
difference operator.  The first two terms preserve the FoundationPose object
projection and WiLoR keypoint reprojection. The contact term draws the hand to 
the object surface, and the temporal term penalizes velocity
and acceleration.  We assign a high weight to $\mathcal L_{\mathrm{FP}}$ because
the object estimates are relatively reliable and smooth in our controlled setting.

\vspace{-1.5mm}
\subsection{Grounding Generated Videos in Simulation with HOI Tracking}
\vspace{-1.5mm}

We use a Sharpa Wave Hand~\cite{sharpa2025wave} mounted on a Franka
Research~3 arm~\cite{franka2022fr3}, simulated with PhysX in Isaac
Gym~\cite{makoviychuk2021isaacgym}.  

\subsubsection{Reference Augmentation and Preprocessing}
We first augment each source trajectory with 5 sampled variations in hand approach and
post-contact object motion:
\vspace{-1.2mm}
\begin{equation}
\begin{gathered}
\begin{aligned}
&\widetilde{\mathbf T}_t^X
  =\mathbf G\mathbf C_t^X\mathbf T_t^X,\quad X\in\{H,O\},\\
&\widetilde{\mathbf J}_t^H
  =(\mathbf G\mathbf C_t^H)\!\cdot\!\mathbf J_t^H,
\end{aligned}\\[-0.2em]
(\mathbf C_t^H,\mathbf C_t^O)
  =\begin{cases}
  (\operatorname{Exp}(\alpha_t\boldsymbol\xi_0),\mathbf I), & t<t_c,\\
  (\operatorname{Exp}(\beta_t\boldsymbol\xi_e),
   \operatorname{Exp}(\beta_t\boldsymbol\xi_e)), & t\geq t_c.
  \end{cases}
\end{gathered}
\label{eq:contact_preserving_augmentation}
\vspace{-1.2mm}
\end{equation}
Here, $t_c$ and $T$ denote first contact and the final frame.
$\mathbf T_t^X$ and $\mathbf J_t^H$ are the source poses and MANO points, and
tildes denote augmented quantities.
$\mathbf C_t^X$ is the perturbation and $\mathbf G$ a trajectory-level yaw
transform.  The exponential map maps sampled twists to $SE(3)$. Before contact,
$\alpha_t=1-t/t_c$ maps $[0,t_c]$ from $1$ to $0$. Afterward,
$\beta_t=(t-t_c)/(T-t_c)$ maps $[t_c,T]$ from $0$ to $1$.  We set both
perturbations to $30\%$ of the source clearance and motion range.  Sharing the
post-contact transform preserves the hand--object relative pose during contact.
Each augmented candidate is then screened for ease of execution. We solve its
FR3 trajectory using inverse kinematics (IK) and slow the whole trajectory with interpolation 
according to defined joint-velocity limits.

\subsubsection{RL Formulation and Training Recipe}
\noindent\textbf{Observation and control.}
We train an asymmetric actor--critic RL to learn the tracking policies. We design the actor input as
\begin{equation}
\mathbf o_t^\pi=[\,\mathbf q_t,\mathbf a_{t-1},\mathbf x_t^w,
\mathbf x_t^{\mathrm{tip}},\widetilde{\mathbf x}_t^o,
\mathbf e_t^{\mathrm{HOI}},\mathbf d_t^{\mathrm{ref}},
\phi(\mathcal M^o)\,],
\label{eq:actor_observation}
\end{equation}
where $\mathbf a_{t-1}$ is the previous raw action, wrist quantities are
expressed in the FR3 base frame, fingertips and the noisy object pose are
expressed relative to the wrist, and $\mathbf e_t^{\mathrm{HOI}}$ contains
current wrist, MANO-keypoint, and object-pose errors.  The remaining terms are
the five reference fingertip--surface distances and a BPS object-shape
encoding.  Inspired by~\cite{li2025maniptrans}, we redundantly encode
hand--object relative geometry such as HOI errors in the observation space. We find that this
simplifies exploration.  The actor receives no joint velocity, force, mass, center of mass,
or retargeted hand-joint target.  The critic additionally receives privileged
simulator state including clean object pose and joint velocity.

The policy is represented as an MLP that outputs $\mathbf a_t=[\mathbf a_t^A,\mathbf a_t^H]\in[-1,1]^{29}$
(seven arm and 22 hand dimensions).  Arm actions are joint deltas about the
measured state, $\bar{\mathbf q}_t^A=\mathbf q_t^A+(0.03\,\mathrm{rad})
\mathbf a_t^A$. Hand actions are mapped through the joint limits to
absolute targets $\bar{\mathbf q}_t^H$.  Both targets are smoothed by an
exponential moving average (EMA),
$\mathbf q_t^{X,\mathrm{PD}}=\alpha_X\bar{\mathbf q}_t^X+
(1-\alpha_X)\mathbf q_{t-1}^{X,\mathrm{PD}}$ with
$(\alpha_A,\alpha_H)=(0.20,0.10)$, The result is
sent to 30-Hz position PD control.

\noindent\textbf{Reward.}
Let $\kappa_\alpha(e)=\exp(-\alpha e)$.  The reward is
\begin{equation}
r_t=r_t^H+r_t^O+r_t^{\mathrm{near}}+r_t^{\mathrm{multi}}
    +r_t^{\mathrm{lift}}+r_t^{\mathrm{eff}}-r_t^{\mathrm{reg}}.
\label{eq:tracking_reward}
\end{equation}
The first two terms track the demonstration.  Inspired by the contact shaping in~\cite{zhang2025robustdexgrasp}, $r_t^{\mathrm{near}}$ and
$r_t^{\mathrm{multi}}$ encourage fingertip approach and multi-finger contact.
Together with action-smoothness and effort regularization, they improve
sim-to-real transfer, while $r_t^{\mathrm{lift}}$ makes demonstrations
containing grasp-and-lift motion easier to learn.

In particular, the imitation terms are
\vspace{-1.2mm}
{\small
\begin{align}
r_t^H=g_t^L\big[&0.1\kappa_{40}(e_{w,p})+3\kappa_{1.91}(e_{w,R})
+\rho_t\!\sum\nolimits_{j\in\mathcal G}w_j\kappa_{\alpha_j}(e_j) \notag\\
&+0.1\kappa_1(e_{w,v})+0.05\kappa_1(e_{w,\omega})
+0.1\kappa_1(e_{J,v})\big],\\
r_t^O=g_t^C\big[&5\kappa_{80}(e_{o,p})+\kappa_3(e_{o,R})+0.1\kappa_1(e_{o,v}) \notag\\
&+0.1\kappa_1(e_{o,\omega})\big].
\label{eq:tracking_reward_terms}
\vspace{-1.2mm}
\end{align}}
A hat denotes a reference, and all translational errors use the FR3 base
frame.  For wrist or object $x\in\{w,o\}$, we use
$e_{x,p}=\|\mathbf p_t^x-\hat{\mathbf p}_t^x\|_2$,
$e_{x,v}=\|\mathbf v_t^x-\hat{\mathbf v}_t^x\|_1/3$, geodesic rotation
error $e_{x,R}$, and
$e_{x,\omega}=\|\boldsymbol\omega_t^x-\hat{\boldsymbol\omega}_t^x\|_1/3$.
For MANO group $\mathcal G_j$, $e_j$ is the mean pointwise $L_2$ position
error and $e_{J,v}$ the mean absolute Cartesian velocity error over all MANO
points.  In the order thumb, index, middle, ring, pinky, level-1, and level-2,
$\mathbf w=(0.9,0.8,0.75,0.6,0.6,0.5,0.3)$ and
$\boldsymbol\alpha=(100,90,80,60,60,50,40)$.  We set $\rho_t=2$ in free
space and $0.75$ during contact.

Let $d_i$ and $F_i$ be elastomer fingertip--object distance and contact force,
$c_t^{\mathrm{ref}}$ the number of reference fingertip contacts, and
$c_t=\sum_i\mathbb I[F_i>0.1\,\mathrm N,\,d_i\leq0.015\,\mathrm m]$ the
rollout contact count. The binary gates are
$g_t^C=\mathbb I[c_t^{\mathrm{ref}}>0\lor c_t>0]$ and
$g_t^L=1-\mathbb I[\hat h_t-h_0>0.05\land h_t-h_0\leq0.05]$.
Thus, $g_t^C$ activates object tracking upon reference or rollout contact,
whereas $g_t^L=0$ suppresses dense hand tracking when a reference lift is not reproduced.
Define
$g_t^{\mathrm{ref},3}=\mathbb I[c_t^{\mathrm{ref}}\geq3]$ and
$g_t^{\mathrm{ref},+}=\mathbb I[c_t^{\mathrm{ref}}>0]$.  The two contact
regularizers are
$r_t^{\mathrm{near}}=0.5g_t^{\mathrm{ref},3}\sum_i
\exp(-([d_i-0.015]_+/0.020)^2)$ and
$r_t^{\mathrm{multi}}=g_t^{\mathrm{ref},+}\{1.25\text{ if }c_t=4;
2.5\text{ if }c_t=5;0\text{ otherwise}\}$.  We further use
$r_t^{\mathrm{lift}}=3\mathbb I[\hat h_t-h_0>0.05]
\mathbb I[h_t-h_0>0.05]$ and
$r_t^{\mathrm{eff}}=0.5\kappa_{10}(P_t)$.
Here $h_t$, $\hat h_t$, and $h_0$ are current, reference, and placed object
heights, and $P_t=\sum_{k\in H}|\tau_{t,k}\dot q_{t,k}|$ is the hand-joint
mechanical power.  Finally, $r_t^{\mathrm{reg}}$ penalizes consecutive
arm/hand raw-action differences with weights $(0.05,0.025)$, joint velocities
normalized by $15\%$ of their limits with weights $(0.10,0.04)$, and normalized
torque above $20\%$ of the effort limit with weight $0.2$.

\noindent\textbf{Policy optimization.}
We optimize with SAPG~\cite{singla2024sapg}, which partitions parallel
rollouts across Proximal Policy Optimization (PPO) agents with different
exploration settings and aggregates their experience into a shared policy update.
We pair this exploration mechanism with contact- and lift-aware rewards to
guide learning from noisy multi-trajectory references.

\noindent\textbf{Domain randomization.}
At every episode, we randomize arm and hand PD gains, object mass, object,
table, and fingertip friction, and per-step actuation latency.  We also
corrupt the actor's object-pose observation while retaining clean simulator
state for the privileged critic.

\noindent\textbf{Episode initialization and termination.}
During training, we sample a reference frame and initialize the arm and object
consistently with it.  We retain the retargeted wrist pose but reset
all 22 hand joints to an open configuration.  Directly using the retargeted
finger pose can initialize hand links inside the object or table, producing
large depenetration impulses and unstable learning.  An episode succeeds
at the end of the reference and terminates early for object/hand tracking
failure or exceeding joint speed thresholds.

\begin{figure*}[t]
    \centering
    \includegraphics[width=\textwidth]{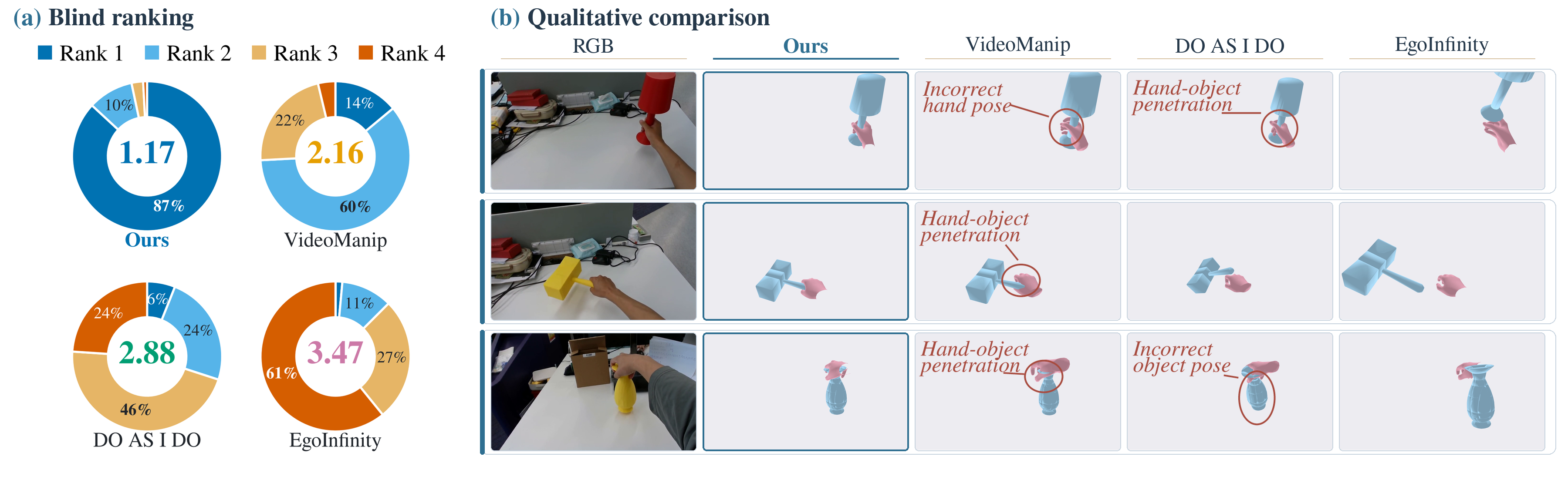}
    \par\vspace{-1.4em}
    \caption{\figtitle{Blind ranking and qualitative comparison on generated videos.}
    (a) Rank distributions across 480 judgments per method; centers report
    mean rank (lower is better). (b) Representative frames using identical
    rendering, where our method better preserves object pose and hand--object
    placement.}
    \label{fig:hoi_blind_qualitative}
    \vspace{-0.7em}
\end{figure*}

\begin{table*}[!t]
    \centering
    \caption{HOI reconstruction benchmark and component ablations.}
    \label{tab:hoi_reconstruction_ablation}
    \label{tab:hoi_gt_benchmark}
    \vspace{-0.4em}
    {\fontsize{7.82}{9.016}\selectfont
    \setlength{\tabcolsep}{2.5pt}
    \renewcommand{\arraystretch}{1.04}
    \begin{tabular*}{\textwidth}{@{\extracolsep{\fill}}lcccccccccc@{}}
        \toprule
        \textbf{Method} &
        \multicolumn{5}{c}{\textbf{H2O (40)} $\cdot$ 150 frames/clip $\cdot$ 1280$\times$720 px} &
        \multicolumn{5}{c}{\textbf{HO-Cap (40)} $\cdot$ 150 frames/clip $\cdot$ 640$\times$480 px} \\[-0.4pt]
        \cmidrule(lr){2-6}\cmidrule(lr){7-11}
        & \textbf{ADD-S}$\downarrow$
        & \textbf{MRRPE}$\downarrow$
        & \textbf{CDev}$\downarrow$
        & \textbf{Avg. Rank}$\downarrow$
        & \textbf{Time}$\downarrow$
        & \textbf{ADD-S}$\downarrow$
        & \textbf{MRRPE}$\downarrow$
        & \textbf{CDev}$\downarrow$
        & \textbf{Avg. Rank}$\downarrow$
        & \textbf{Time}$\downarrow$ \\
        \midrule
        \textbf{Ours}
        & \textcolor{tblBest}{\textbf{4.82}} & \textcolor{tblSecond}{\textbf{84.82}} & \textcolor{tblSecond}{\textbf{91.29}}
        & \textcolor{tblBest}{\textbf{1.67}} & 12.33
        & \textcolor{tblBest}{\textbf{3.88}} & \textcolor{tblSecond}{\textbf{54.87}} & \textcolor{tblSecond}{\textbf{41.36}}
        & \textcolor{tblBest}{\textbf{1.67}} & 6.73 \\
        \textcolor{tblGray}{\hspace{0.35em}w/o depth align.}
        & 5.14 & 86.97 & \textcolor{tblBest}{\textbf{85.97}} & \textcolor{tblSecond}{\textbf{2.33}} & --
        & 6.01 & 57.04 & \textcolor{tblBest}{\textbf{39.34}} & \textcolor{tblSecond}{\textbf{2.67}} & -- \\
        \textcolor{tblGray}{\hspace{0.35em}w/o joint opt.}
        & \textcolor{tblSecond}{\textbf{5.10}} & 127.56 & 143.54 & 3.67 & --
        & \textcolor{tblSecond}{\textbf{4.02}} & 102.08 & 112.43 & 4.33 & -- \\
        \addlinespace[1pt]
        VideoManip
        & 11.62 & 166.20 & 184.89 & 6.00 & \textcolor{tblBest}{\textbf{5.97}}
        & 5.09 & 84.75 & 100.59 & 4.00 & \textcolor{tblBest}{\textbf{4.72}} \\
        EgoInfinity
        & 5.47 & 102.96 & 149.00 & 4.33 & \textcolor{tblSecond}{\textbf{11.12}}
        & 7.73 & 81.52 & 118.32 & 5.00 & \textcolor{tblSecond}{\textbf{5.79}} \\
        DO AS I DO
        & 5.52 & \textcolor{tblBest}{\textbf{71.83}} & 95.88 & 3.00 & 114.93
        & 14.50 & \textcolor{tblBest}{\textbf{46.53}} & 78.96 & 3.33 & 112.40 \\
        \bottomrule
    \end{tabular*}}
    \vspace{-1.8em}
\end{table*}

\subsubsection{Policy Distillation}

Direct training on the full reference set remains difficult under limited
compute, whereas per-trajectory training~\cite{liu2025dextrack} scales poorly.
We therefore train at most two multi-skill experts per object category, each
covering about 40 source trajectories (roughly 200 after augmentation). Each
expert is obtained either by training from scratch on all trajectories of that
category or, when this is less effective, by training on the full multi-object
set and then fine-tuning on the target category. We distill these experts into a single
policy through behavior cloning, then fine-tune the distilled policy with
DAgger~\cite{ross2011reduction} to mitigate distribution shift.

\section{EXPERIMENTS}
\vspace{-1.0mm}
\subsection{Ablation and Benchmark for HOI Reconstruction}
\vspace{-1.0mm}
\noindent\textbf{Setup.}
We evaluate reconstruction in two settings. For generated videos without 3-D
ground truth, four reviewers rank anonymized outputs of our method and three
baselines on 120 clips sampled uniformly from 2,500 generations. Ties are
allowed and method order is randomized, yielding 480 rankings per method. For
quantitative evaluation, we use 40 fixed-view RGB-D clips from each
of H2O~\cite{kwon2021h2o} and HO-Cap~\cite{wang2025hocap}, both with 3-D annotations.
We compare the full method, its ablations, and the baselines using three metrics.

Let $\widehat{T}_t,T_t^*$ denote the predicted and ground-truth object-to-camera
transforms and $\widehat{\mathbf w}_t,\mathbf w_t^*$ the corresponding wrist roots.
ADD-S~\cite{xiang2018posecnn} measures closest-point object alignment as
$\mathrm{ADD\mbox{-}S}_t=|\mathcal V|^{-1}\sum_{\mathbf v\in\mathcal V}
\min_{\mathbf u\in\mathcal V}\|\widehat{T}_t(\mathbf v)-T_t^*(\mathbf u)\|_2$,
where $\mathcal V$ contains at most 500 sampled CAD vertices. MRRPE$_{ro}$~\cite{fan2023arctic}
measures error in the wrist-to-object-center vector,
$\mathrm{MRRPE}_{ro,t}=\|(\widehat{\mathbf w}_t-\widehat{T}_t(\bar{\mathbf v}))-
(\mathbf w_t^*-T_t^*(\bar{\mathbf v}))\|_2$, where $\bar{\mathbf v}$ is the mesh
centroid. CDev measures preservation of ground-truth contacts: we match each
MANO vertex to its nearest CAD vertex, retain pairs within 3~mm, and average
their distances under the predicted hand and object poses. We report ADD-S in
centimeters and MRRPE$_{ro}$/CDev in millimeters. Avg. Rank averages the three metric ranks and excludes runtime on a single RTX 5880 (minutes/clip).

\noindent\textbf{Baselines and ablations.}
We compare three baselines. \emph{VideoManip}~\cite{chen2026videomanip}
reconstructs metric hand--object trajectories from monocular video and refines
the hand pose using predicted contacts. \emph{EgoInfinity}~\cite{wang2026egoinfinity}
calibrates hand and object estimates into a shared metric frame and refines
object trajectories using inferred interaction states. \emph{DO AS I
DO}~\cite{paliwal2026doasido} combines world-space hand tracking with
SAM3-based \cite{sam3dteam2025sam3d} object tracking and depth-based hand--object alignment.  In ablation studies, \emph{w/o depth align.}
passes raw known-intrinsics MoGe-2 depth to FoundationPose and the joint optimizer,
while \emph{w/o joint opt.} removes Joint Hand--Object Optimization
(Eq.~\eqref{eq:hoi_reconstruction}) and retains the independent FoundationPose
object and WiLoR MANO estimates. All other components remain unchanged.
Because \emph{w/o depth align.} uses no measured depth, it also provides a
comparison matched to the monocular baselines with respect to sensing modality;
all methods are adapted to use the shared object mesh.

\noindent\textbf{Results.}
Human raters place our method first in 417 of 480 judgments (87\%), with a mean
rank of 1.17 versus 2.16 for VideoManip, 2.88 for DO AS I DO, and 3.47 for
EgoInfinity (Fig.~\ref{fig:hoi_blind_qualitative}(a)). The qualitative examples
likewise preserve the observed object orientation and grasp placement, whereas
the baselines show displacement, detachment, or hand--object penetration
(Fig.~\ref{fig:hoi_blind_qualitative}(b)). Across H2O and HO-Cap, our method
achieves the best Avg. Rank (1.67), the lowest ADD-S (4.82~cm on H2O and
3.88~cm on HO-Cap), and second-best relative-geometry metrics
(Table~\ref{tab:hoi_gt_benchmark}).
DO AS I DO uses the reconstructed hand as a metric anchor for per-frame object
translation~\cite{paliwal2026doasido}. MRRPE$_{ro}$ naturally favors this
hand-referenced alignment because it measures only wrist-to-object displacement
and cancels absolute-depth errors shared by the hand and object. Its higher
ADD-S and CDev, together with the human evaluation, show that this structural
advantage under MRRPE$_{ro}$ does not imply better overall 3-D quality.

The ablations separate the roles of the two reconstruction stages. Even without
depth alignment, our method has a better Avg. Rank than every baseline on both
H2O (2.33) and HO-Cap (2.67). Its main degradation relative to the full method
is in ADD-S, particularly on HO-Cap, showing that the overall quantitative gain
does not arise solely from the measured depth anchor. Joint optimization
markedly improves MRRPE$_{ro}$ and CDev on both datasets, confirming better
relative hand--object geometry while preserving absolute object placement.

\noindent\textbf{\emph{Remark:} Usable reference yield.}
Of the 2,500 generated clips, 83\% passed the reconstruction quality check,
yielding about 2,000 usable references. As a point of reference, DO AS I DO's
audit of 2,000 in-the-wild 100 Days of Hands clips found meaningful HOI in 187
(9.4\%) and reconstructable HOI in 83 (4.2\%)~\cite{paliwal2026doasido}. The
criteria differ and the two sources are complementary, but the rates suggest
that image-and-language conditioning can reduce human effort required to
screen out irrelevant or unreconstructable videos.

\begin{figure}[t]
  \centering
  \includegraphics[width=\columnwidth]{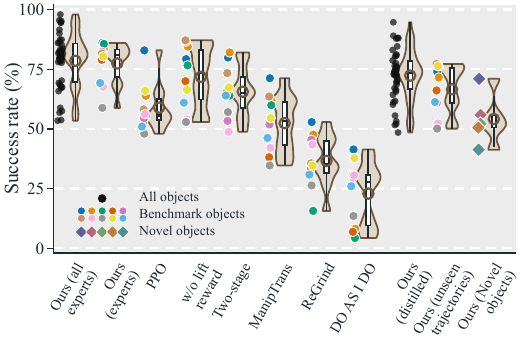}
  \par\vspace{-1em}
  \caption{\figtitle{Success rate statistics.} Black circles, colored circles, and
    diamonds denote 42 training, ten benchmark, and five novel objects,
    respectively. Violins show distributions; boxes, bars, whiskers, and hollow
    circles mark interquartile ranges, medians, $1.5\times$IQR, and object-level
    means. Higher
    is better.}
  \label{fig:ablation_results}
  \vspace{-1em}
\end{figure}

\begin{figure}[t]
  \centering
  \includegraphics[width=\columnwidth,trim=0 2.5pt 0 0,clip]{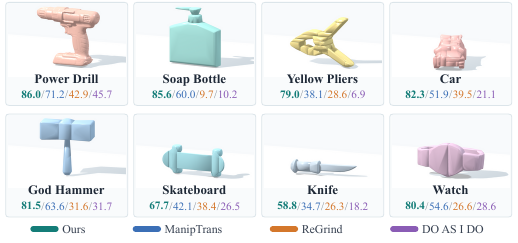}
  \par\vspace{-0.8em}
  \caption{\figtitle{Examples of manipulated objects.} Values are the per-object success rates
    (\%) of the methods in Fig.~\ref{fig:ablation_results}, in the same order. Meshes are
    normalized for visualization.}
  \label{fig:example_objects}
\end{figure}

\vspace{-2mm}
\subsection{Ablation and Benchmark for RL Formulation and Optimization}
\label{sec:rl_benchmark}
\vspace{-1.5mm}
\noindent\textbf{Setup.}
Fig.~\ref{fig:ablation_results} reports the expert policies on all 42 training
objects and a 10-object benchmark, together with the ablations, baselines, and
distilled policy. The latter is evaluated on all training objects, 300 unseen
trajectories of the benchmark objects, and five novel objects. Some
benchmark objects are illustrated in Fig.~\ref{fig:example_objects}.
Benchmark methods share the evaluation reference set, with matched training
budgets for learned policies.
For each method--object experiment, we
report success rate from 20,000 evaluation rollouts, each starting from the
first reference frame.
A trajectory is successful if its rollout reaches the final reference without exceeding a $4\,$cm object-position error, 
a $30^\circ$ object-rotation error, or hand-keypoint thresholds of $6/6/8/10/12/
12\,$cm for the thumb/index/middle/ring--pinky/level-1/level-2 groups.
Tracking errors are averaged over \emph{intent-executing} segments, i.e., after contact and a
$5\,$cm lift until the first drop below $2.5\,$cm for grasp-and-move/pose-adjustment trajectories, and
after first contact for push-and-pull trajectories, with segments shorter than eight steps discarded.
Across all operation types, we report object-position error $\|\mathbf p_t-\hat{\mathbf p}_t\|_2$,
geodesic rotation error $2\arccos(|\mathbf q_t^\top\hat{\mathbf q}_t|)$, and mean
Euclidean error over 27 non-wrist hand keypoints.

\begin{table}[t]
  \centering
  \caption{Tracking errors over intent-executing trajectories.}
  \label{tab:intent_tracking_errors}
  \vspace{-0.6em}
  {\fontsize{7.82}{9.016}\selectfont
  \setlength{\tabcolsep}{5pt}
  \begin{tabular}{lccc}
    \hline
    Method & Pos. (mm) $\downarrow$ & Rot. ($^\circ$) $\downarrow$ & Hand (mm) $\downarrow$ \\
    \hline
    Ours       & \textbf{10.2} & 17.4 & \textbf{36.2} \\
    PPO        & 11.6 & 26.4 & 39.3 \\
    w/o lift reward    & 11.4 & \textbf{13.3} & 39.1 \\
    Two-stage   & 23.6 & 25.8 & 42.6 \\
    ManipTrans        & 25.8 & 23.6 & 39.7 \\
    ReGrind           & 17.7 & 14.0 & 76.8 \\
    \multicolumn{4}{@{}c@{}}{\leaders\hbox{\raisebox{1.2ex}[0pt][0pt]{\rule{2pt}{0.35pt}}\hskip2pt}\hfill\kern0pt}\\[-1.4ex]
    DO AS I DO (floating)   & 35.3 &  7.1 & 36.9 \\
    \hline
    \end{tabular}}
\end{table}

\begin{figure*}[!t]
  \centering
  \includegraphics[width=\textwidth]{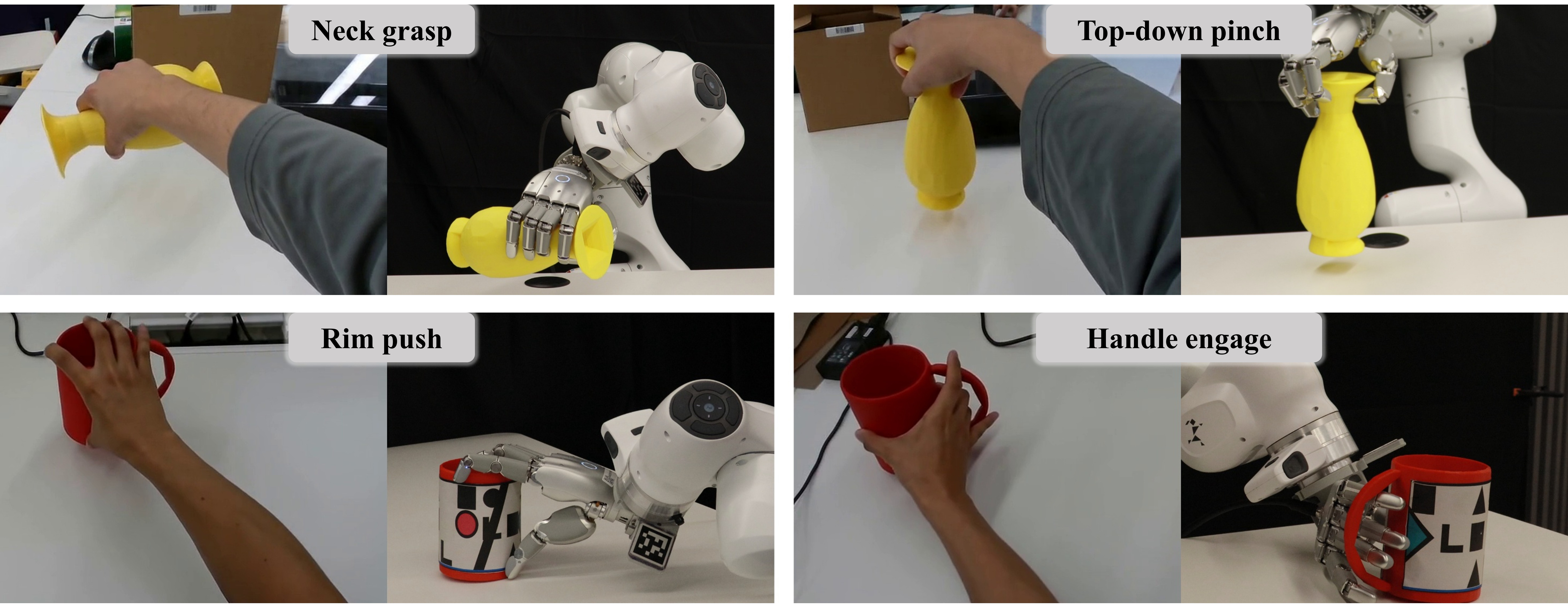}
  \par\vspace{-0.5em}
  \caption{\figtitle{Generated human references and corresponding real-world executions.}
  All shown reference trajectories are unseen during policy training. The evaluated
  tasks comprise jar-neck pose adjustment, jar top-down grasp-and-move, mug-rim
  pushing, and mug-handle pulling.}
  \label{fig:real_world_results}
\end{figure*}

\noindent\textbf{Baselines and ablations.}
We compare with: 1) \emph{PPO}, which replaces SAPG while keeping the remaining
formulation of our method; 2) \emph{w/o lift reward}, which removes
$r_t^{\mathrm{lift}}$; 3) \emph{Two-stage}, our method with a hand imitation stage followed by a residual policy, which is similar to ManipTrans~\cite{li2025maniptrans}; 
4) \emph{ManipTrans}, a two-stage
method that learns hand-only trajectory imitation followed by a residual
policy; 5) \emph{ReGrind}~\cite{feng2026regrind}, which uses interaction-aware
retargeting to provide nominal targets for a residual PPO policy with
object-keypoint and wrist tracking; and 6) \emph{DO AS I
DO}~\cite{paliwal2026doasido}, a per-reference MPPI-style dynamics-aware
retargeting method that directly optimizes an (open-loop) action sequence.  The residual
stages of ManipTrans and the ReGrind policy are originally optimized per
reference but can be adapted to our multi-trajectory setting.  We adapt both to our arm--hand embodiment and dynamics (including domain randomization) and 
align their observation spaces
with ours. For DO AS I DO, we retain its floating-wrist formulation and align some of the dynamics parameters 
like hand PD gains and joint limits.  

\noindent\textbf{Results.}
Across all 42 training objects, our expert policies reach a $78.6\%$ macro mean and
$80.5\%$ median (Fig.~\ref{fig:ablation_results}). 
Distillation reduces the macro mean and median by only $4.2$ and $6.0$
percentage points, respectively. The distilled controller further reaches
$66.6/68.8\%$ and $54.2/52.2\%$ mean/median on unseen reference trajectories and
novel objects, respectively, showing transfer without retraining despite the
substantial variation in manipulation difficulty.
On the 10-object benchmark, Ours (experts)
achieves the highest mean and median, $77.4\%$ and $81.0\%$.  Replacing SAPG with
PPO, removing the lift reward, or switching to two-stage training reduces
success on $10/10$, $7/10$, and $10/10$ objects, respectively.  We sometimes
observe faster early learning with two-stage training, but hypothesize that,
when reconstructed hand--object relations are noisy, its first stage can impose
suboptimal hand trajectories that restrict later-stage RL exploration. 
We note that ManipTrans was instead developed with cleaner motion-capture 
data~\cite{li2025maniptrans}, which explains this design choice.  Our
two-stage variant nevertheless exceeds ManipTrans on every benchmark object by
$13.2$ points on average, indicating that the reward formulation and
optimizer together improve performance under the same training structure.
Together with the lift-reward and two-stage
ablations that retain SAPG, these results support the joint choice of reward formulation, training
structure, and optimizer for noisy multi-trajectory tracking.
ReGrind's lower success may arise because interaction-aware
retargeting propagates reconstruction noise into distorted nominal targets,
as well as its simpler observation and reward design provide less guidance for RL exploration.  

Table~\ref{tab:intent_tracking_errors} reports tracking errors computed 
over intent-executing segments.  Our method achieves the lowest object-position
and hand-keypoint errors among the arm--hand methods.  
We note that DO AS I DO reports a lower rotation error, while this number is
not directly comparable since we find that DO AS I DO can not successfully 
plan a large number of pose-adjust trajectories that are excluded as not meeting the intent-executing criteria. 
\vspace{-1mm}
\subsection{Real-World Experiments}
\vspace{-1mm}
\noindent\textbf{Real-World Setup and Onboard Perception.}
An Intel RealSense D455 observes
the workspace from a viewpoint largely free of arm occlusion and provides
online object poses through FoundationPose~\cite{wen2024foundationpose}. Another
Intel RealSense D435 is used only to capture the scene image that conditions
video generation, which is removed before execution to avoid collision with the robot
and does not participate in closed-loop perception. Because deployment is sensitive to translational calibration
error, beyond typical hand-eye calibration, we command the robot rigidly grasp an object and use the known end-effector pose to estimate
a shared constant bias in the FoundationPose translation across objects, analogous to the Tsai--Lenz calibration method.

\begin{table}[t]
  \vspace{0.5em}
  \centering
  \caption{Real-world success on unseen trajectories (10 trials/task).}
  \label{tab:real_world_success}
  {\footnotesize
  \setlength{\tabcolsep}{0pt}
  \renewcommand{\arraystretch}{0.90}
  \begin{tabular*}{\columnwidth}{@{\extracolsep{\fill}}lccccc@{}}
    \toprule
    Task & Jar neck & Jar top & Mug rim & Mug handle & Overall \\
    Type & Pose-adjust & Grasp-and-move & Push-and-pull & Push-and-pull & -- \\
    Success & 6/10 & 6/10 & 7/10 & 8/10 & \textbf{27/40} \\
    \bottomrule
  \end{tabular*}}
  \vspace{-1.3em}
\end{table}

\noindent\textbf{Results.}
Here we present the results of evaluating the distilled controller on 40 unseen video plans. The
four tasks are jar-neck pose adjustment, jar top-down grasp-and-move, mug-rim
pushing, and mug-handle pulling, with each task having 10 trials. Each plan is conditioned on a scene image and a
new language instruction. 
Success follows the same trajectory-completion criteria and tracking-error
thresholds as in Sec.~\ref{sec:rl_benchmark}, using object poses from
FoundationPose and hand keypoints computed by forward kinematics (FK) from
measured robot joint positions. The controller achieves 27/40 successes overall,
with per-task results reported in Table~\ref{tab:real_world_success}.
Representative failures are discussed in
Sec.~\ref{sec:discussion}.
As shown in Fig.~\ref{fig:real_world_results}, the jar-neck task uses a
small-diameter power grasp that leaves the opening accessible, whereas the
top-down task uses a multi-finger pinch at the opening. These grasp choices are
part of the interaction specified by the video plan, highlighting the importance
of retaining both hand and object motion in the reference compared to object-only ones \cite{kuang2026dex4d}.

\section{DISCUSSION AND CONCLUSION}
\label{sec:discussion}
\subsubsection{Failure Modes and Potential Improvements}

\noindent\textbf{Training-time failures.}
Some generated interactions appear plausible but are difficult for a robotic
hand.  For example, humans can lift large objects through palm friction far
from force closure, whereas such motions are hard to learn in simulation.  Our
method and the baselines also struggle with flat or thin objects when the grasp
lies near the table.  Task-specific RL policies~\cite{zhang2025robustdexgrasp}
may provide useful action supervision or motion priors \cite{luo2026sonic} to address these issues.

\noindent\textbf{Deployment failures.}
Contact transitions are sensitive to the sim-to-real gap and can cause large
tracking deviations.  Although the controller shows basic recovery and retry
behavior, it often fails after leaving the reference.  Embedding training in a more generalized
goal-conditioned MDP beyond merely tracking could provide more diverse recovery data.

\subsubsection{Towards Versatile Dexterous Controllers}

Our skill set does not systematically cover in-hand manipulation, which appears
only incidentally in generated videos.  Video references are less reliable for
such contact-rich motion because reconstruction noise obscures subtle
finger--object motion.  We therefore aim to combine multiple data sources in a
unified controller for everyday manipulation of single rigid objects. Nevertheless, the
grasping and non-prehensile skills learned here  demonstrate the
potential of generated video and mark a concrete step toward versatile
dexterous control.

\section*{ACKNOWLEDGMENT}

We thank Kaihan Chen for assistance with implementing the baseline methods, and
Ruoqu Chen and Kechun Xu for early suggestions on key technical design choices.  We also
thank Junxiao Lin, Yipeng Pan, Xingyu Ji, and Mingjie Zhou for conducting the
human preference evaluation of our reconstruction pipeline.
\vspace{-0.1cm}

\bibliographystyle{ieeetr}
\bibliography{references}

\end{document}